\pdfoutput=1

\documentclass[11pt]{article}

\usepackage{acl}

\usepackage{times}
\usepackage{latexsym}
\usepackage{hyperref}
\usepackage{comment}
\usepackage{booktabs}
\usepackage[T1]{fontenc}
\usepackage{todonotes}

\usepackage[utf8]{inputenc}

\usepackage{microtype}
\usepackage{graphicx}
\usepackage{algorithm}
\usepackage{algpseudocode}
\usepackage{amsmath}
\usepackage{amssymb}
\usepackage{makecell}

\usepackage{inconsolata}

\title{SSP: Self-Supervised Prompting for Cross-Lingual Transfer to Low-Resource Languages using Large Language Models}

\author{Vipul Rathore \hskip 1em  
Aniruddha Deb  \hskip 1em
Ankish Chandresh  \hskip 1em
Parag Singla \hskip 1em Mausam \\
        Indian Institute of Technology \\
  New Delhi, India \\  
   \texttt{rathorevipul28@gmail.com},  \texttt{aniruddha.deb.2002@gmail.com} \\ 
   \texttt{iitdelhi24ankish@gmail.com}, 
   \texttt{parags@cse.iitd.ac.in},  \texttt{mausam@cse.iitd.ac.in}}

\begin{document}
\maketitle
\begin{abstract}
Recently, very large language models (LLMs) have shown exceptional performance on several English NLP tasks with just in-context learning (ICL), but their utility in other languages is still underexplored. We investigate their effectiveness for NLP tasks in low-resource languages (LRLs), especially in the setting of \textit{zero-labelled} cross-lingual transfer (0-CLT), where no labelled training data for the target language is available -- however training data from one or more related medium-resource languages (MRLs) is utilized, alongside the available unlabeled test data for a target language.  We introduce Self-Supervised Prompting (SSP), a novel ICL approach tailored for the 0-CLT setting.

SSP is based on the key observation that LLMs output more accurate labels if in-context exemplars are from the target language (even if their labels are slightly noisy). To operationalize this, since target language training data is not available in 0-CLT, SSP operates in two stages. In Stage I, using source MRL training data, target language's test data is noisily labeled. In Stage II, these noisy test data points are used as exemplars in ICL for further improved labelling. Additionally, our implementation of SSP uses a novel Integer Linear Programming (ILP)-based exemplar selection that balances similarity, prediction confidence (when available) and label coverage. Experiments on three tasks and eleven LRLs (from three regions) demonstrate that SSP strongly outperforms existing SOTA fine-tuned and prompting-based baselines in 0-CLT setup.

\end{abstract}

\section{Introduction}
Very large language models (LLMs) such as GPT-3.5-Turbo \& GPT-4 \cite{ouyang-etal-2022-impact, achiam2023gpt} show exceptional performance on a variety of NLP and reasoning tasks via \textit{In-Context Learning} (ICL) \cite{brown2020language,chowdhery2022palm}. ICL feeds a task-specific instruction along with a few exemplars, appended with the test input, to the LLM. As LLMs can be highly sensitive to exemplars \cite{zhao2021calibrate}, exemplar retrieval is crucial for ICL. 


While LLMs have shown excellent performance on English tasks,  their utility on other languages is relatively underexplored. In this work, we study \textit{zero-labelled cross-lingual transfer} (0-CLT) to low-resource languages (LRLs) -- a setting where labeled task data from one or more related medium-resource languages (MRLs) is available, but no labeled data exists for the target LRL. We also additionally leverage the available test sentences (unlabeled) of the target language. This is in contrast to \cite{wan-etal-2023-better, wan-etal-2023-universal}, who utilize a set of external unlabelled sentences for English tasks and pose this as a \emph{transductive} zero-shot setting. The high cost of annotating LRL sentences for new tasks or domains underscores the relevance of the 0-CLT setting for non-English languages.

Cross-lingual transfer has been addressed through standard fine-tuning \cite{muller2021being, alabi2022adapting}, and language adapters \cite{pfeiffer2020mad, ustun2020udapter, rathore2023zgul}, but there is limited work on cross-lingual ICL. There are two exceptions \cite{ahuja2023mega, asai2023buffet}, where ICL is employed with exemplars from a source language, but they use uniformly  random sampling for exemplar selection, resulting in performance inferior to cross-lingually fine-tuned models, such as mBERT and XLM-R \cite{Devlin2019BERTPO, conneau2020unsupervised}.

\begin{figure}
    \centering
    \includegraphics[width=\columnwidth, height = 5cm]{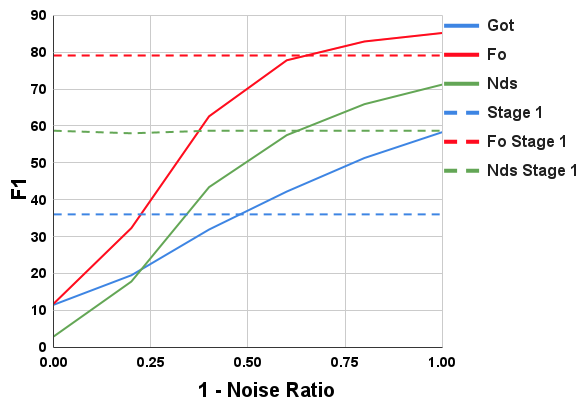}
    \caption{Llama2 70B, prompted with target LRL exemplars, along with artificially injected label noise (x-axis). Dashed lines represent performances when prompted with source MRL exemplars.}
    \label{fig:varynoise}
    \vspace*{-1ex}
\end{figure}
In our preliminary experiments, we prompt the Llama2-70B model with exemplars from source MRLs, and compare its performance with the same LLM prompted with exemplars from the target LRL. We vary the label noise on the target exemplars. Unsurprisingly, LLMs show better performance with less label noise. More interestingly, we find that a reasonably-sized noise region exists (see Figure \ref{fig:varynoise}), such that if the exemplar noise is within that range, then the overall performance is higher than prompting with accurate source language data. 

\begin{figure*}
    \centering
    \includegraphics[scale=0.73]{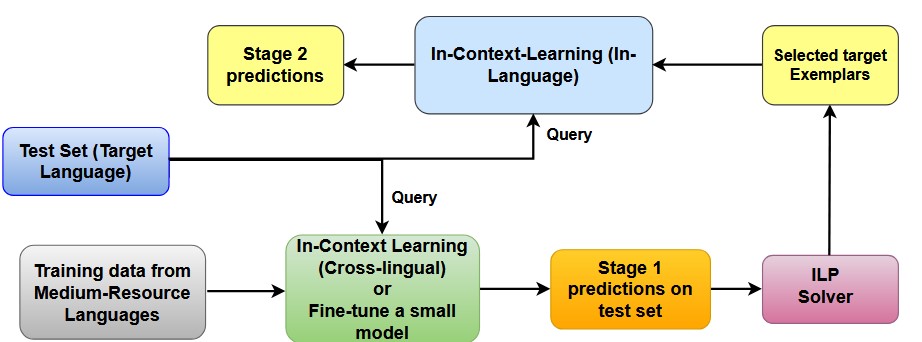}
    \caption{SSP Paradigm for Cross-Lingual Transfer to target low-resource language}
    \label{fig:pipeline}
\end{figure*}

Armed with this observation, we present Self-Supervised Prompting (SSP) -- a novel ICL framework for 0-CLT to LRLs. Since the target LRL training data is not available in 0-CLT, SSP operates in two stages. In Stage I, SSP labels all test instances of LRL using training data from MRL. This may be done by LLM prompting (as in the experiment above), or using any other existing approaches for 0-CLT, such as by fine-tuning or adapters. Once (noisy) labels on target LRL are obtained, in Stage II, SSP uses ICL using these noisy test data points (except itself) as exemplars for further performance improvement.
Additionally, to select the best exemplars, we develop a novel Integer Linear Programming (ILP) based selection approach, which balances the various objectives of (1) similarity of exemplar with test sentence, (2) high confidence in label predictions, and (3) coverage of the various labels for better task understanding. Figure \ref{fig:pipeline} gives an overview of our proposed pipeline.

We define 3 scenarios for our zero-labelled setup - (1) 0-CLT: Only the available test sentences of the target language are used, with no additional unlabelled data, (2) 0-CLT-U: the full wikipedia data available for target language is utilized, and (3) 0-CLT-T: a translation model supporting the target language is leveraged. The primary focus of this work is on 0-CLT (setting 1). However, we also conduct stage 1 experiments for both 0-CLT-U and 0-CLT-T settings. This enables us to comprehensively assess SSP's effectiveness across varying degrees of noise in stage I.

We perform experiments on sequence labelling tasks (POS and NER), and natural language inference (NLI) -- a text classification task. Our datasets encompass eleven low-resource languages from typologically diverse language families and three regions: African, Germanic and American. Our experiments show consistent and substantial improvements over existing fine-tuning as well as simpler ICL-based approaches. We will make both our codebase and prompts publicly accessible.


Our contributions are summarized as follows:
\begin{enumerate}
    \setlength{\itemsep}{-0.5em}
    \vspace{-1.0em}
    \item We Investigate ICL strategies for zero-labelled cross-lingual transfer to LRLs, using labeled data from related MRLs and unlabeled test data from the target language.\
    \item We propose SSP, a two-stage self-supervised prompting paradigm for this task, where the first stage may be done by an LLM or other cross-lingually fine-tuned models. 
    \item We introduce an exemplar selection approach utilizing an ILP. The ILP incorporates similarity to test input along with confidence of prediction (when available), and enforces label coverage constraints for better selection.
    \item Experiments on 3 tasks and 11 languages show that SSP outperforms existing fine-tuning and SOTA LLM-based models in 0-CLT, 0-CLT-U (full unlabeled) as well as 0-CLT-T (translation-based) settings, hence improving labelling in the second iteration, irrespective of the initial labelling method.
\end{enumerate}

\section{Related Work}
An ICL prompt consists of (1) task description: to facilitate the understanding of task, (2) labeled input-output pairs: Written sequentially in order of their relevance to input query, and (3) input itself. 

\noindent
\textbf{Cross-lingual ICL}:
In general, cross-lingual ICL has not been systematically explored in literature. In existing works, prompting is primarily done in a high-resource language, typically English. This is called \textit{cross-lingual (CL) prompting}. This differs from \textit{in-language (IL) prompting}, where examples are retrieved from the candidate pool of the target language itself. This assumes the availability of labeled data for target LRL, which is not true in our zero-labelled (0-CLT) setting.
In response, we develop novel techniques making use of both CL prompting and IL prompting, while not utilizing the gold labels during IL prompting stage. 

Most existing cross-lingual ICL methods use uniformly random input-output pairs for exemplar selection \cite{zhang2021differentiable,winata2021language, ahuja2023mega, asai2023buffet}. Recent approaches \cite{Agrawal2022IncontextES,tanwar-etal-2023-multilingual} address this gap by utilizing \textit{semantic similarity} for cross-lingual retrieval from a high-resource language's labeled data, given the target LRL's instance as query. This is facilitated by embedding-based multilingual retrievers such as multilingual sentence-transformers \cite{reimers2020making}. More recently, OpenAI-based embeddings such as Ada-002 \footnote{\href{https://platform.openai.com/docs/guides/embeddings/embedding-model}{https://platform.openai.com/docs/guides/embeddings/}} have been used effectively for cross-lingual retrieval \cite{nambi2023breaking}. We extend this line of work by also incorporating label confidence and label coverage in exemplar selection. 

\vspace{0.5ex}
\noindent
\textbf{Fine-tuning approaches for Cross-lingual Transfer:} Most approaches rely on fine-tuning a Pretrained LM (PLM) such as BERT or XLM-R on one or more source languages (\cite{muller2021being, alabi2022adapting}) and deploying on an unseen target language. Recently, Language-Adapter based approaches have been found more effective \cite{ustun2020udapter} for cross-lingual transfer settings. For sequence labelling tasks (NER and POS tagging), ZGUL \cite{rathore2023zgul} is a recent SOTA method that leverages ensembling Language Adapters from multiple MRLs to label each word in a target language. We leverage this in our proposed SSP pipeline. \newline
\textbf{Cross-lingual label-projection techniques:} Recent methods \cite{chen2023frustratingly, garcia2023t, le2024constrained} utilize an off-the-shelf translation model \cite{nllb2022} for label-projection in 2 ways -- (1) \emph{Translate-train}: translate from English to target language (X) to generate training data in X, or (2) \emph{Translate-test}: translate test data in X to English to perform label-projection and obtain annotations in X. Although our focus is 0-CLT transfer, we also experiment with these translation models in Stage I, to assess the robustness of SSP across multiple settings.


\section{Self-Supervised Prompting}
We define the setting of zero-labelled cross-lingual transfer (0-CLT) as follows. We are given source training data for a specific task: $D=\{(x_i,lg_i,y_i)\}$, where $x_i$ is the input text in language $lg_i$, and the output is $y_i$. We are additionally given a set of unlabeled test data points $T=\{q_j\}$ from a target language $lg_t$. Our goal is to train a model/create a protocol, using $D$, $T$ and a large pre-trained LLM, that outputs good  predictions on $T$ for the task, assuming that $lg_t$ is a low-resource language, due to which its training data is not available, and that languages $lg_i$ are related to $lg_t$.

Our solution approach, Self-Supervised Prompting (SSP), comprises two key stages as follows. In Stage I, it proposes a noisy labelling for all data points in $T$ using source data $D$. This may be done in different ways, as described next. In Stage II, it uses the LLM and noisy labelling on $T$ from Stage I as exemplars to improve the labellings. Furthermore, SSP uses a novel integer-linear programming based exemplar selection. We now describe each component of our system.

\subsection{Stage I: initial labelling using source data}
\label{subsubsec:sim}
To create a first labelling for all test points, SSP can use any existing approaches for 0-CLT, such as fine-tuning a multilingual language model for the task, or use of language adapters or using our LLM with in-context exemplars from source language. In our experiments, we experiment with adapters and ICL, which we briefly describe next.

\vspace{0.5ex}
\noindent
\textbf{Cross-Lingual ICL: } In the method, we use ICL over LLM for obtaining Stage I labellings. First, we retrieve a set of top-$K$  exemplars from $D$ using each test instance $q_j$ as query. This selection is based on cosine similarity between their \textit{Ada-002} embeddings. The selected exemplars are arranged in descending order of similarity scores, and included in the prompt between the task description (TD) and the input test instance. This approach has two drawbacks. First, since the LLM will typically be a large expensive model -- this will require an LLM call per test data point in Stage I. Second, generally, these LLMs do not expose their logits, hence, we will not have access to prediction confidences from Stage I labellings. 

\vspace{0.5ex}
\noindent
\textbf{Training smaller model(s) using $D$: }
Another possibility is to fine-tune a smaller multilingual LM, such as mBERT or mDeBerta-v3 \cite{he2021deberta} on $D$ for NLI task. For sequence labelling, we can use ZGUL \cite{rathore2023zgul}, which trains source language adapters using $D$, and uses inference-time fusion of source adapters for labelling test data points. These approaches can provide Stage I labellings for $T$ along with prediction confidences, without making any expensive LLM calls.

\subsection{Stage II: in-language ICL using ILP-based exemplar selection}
\label{subsec:ilp}
After Stage I predictions for target instances $T$ are obtained, SSP prompts the LLM to label each test data point $q \in T$, but uses in-context exemplars in target language using Stage I labellings. For exemplar selection, SSP implements a novel integer linear program (ILP) that balances \emph{semantic similarity, prediction confidence} (when available) and \emph{label coverage}. 

Our primary objective is to maximize the aggregated semantic similarity of the selected exemplars, which is obtained using cosine similarity score between their OpenAI Ada-v2 embeddings. In addition, we impose two constraints:
\begin{itemize}
    \item \textbf{Label Coverage}: The ILP tries to ensure the coverage of all labels for the given task in the selected exemplars -- this has been found effective for ICL \cite{min2022rethinking}.
    \item \textbf{Confidence}: In case Stage I predictions are made by a model whose logits are accessible (unlike the OpenAI LLMs), the ILP prefers selection of more confident exemplars. Our hypothesis is that confident predictions are also accurate (assuming the model is well-calibrated), and previous work has shown that performance of LLMs can be sensitive to correctness of exemplars \cite{wei2023larger}
\end{itemize}
SSP formulates these three factors into an ILP as follows. For a dataset $D$ with $n$ examples indexed from $\mathcal{I} = \{1 \ldots n\}$, given a test data point $q_j$, let $z_i$ be a binary variable denoting whether $i^{th}$ test instance $q_i$ is selected as an exemplar. We use a semantic similarity function $\text{sim}(q_i, q_j)$ to get the similarity between two examples. $K$ is the number of exemplars to be selected. Since $q_j$ cannot be an exemplar for itself, we select exemplars from $\mathcal{I} \setminus \{j\}$ only.

Let the set of all labels in the task be $\mathcal{L}$, and the multiset of all labels predicted (using argmax) for example $q_i$ be $L_i$. The Stage I prediction confidence for label $l$ in $q_i$ is denoted as $\hat{y}^i_l$. This confidence is computed as average of probability scores across all predictions of label $l$ in $i^{th}$ sentence (details in Appendix \ref{sec:implement}). The ILP uses a threshold $\tau_l$ for prediction confidence for a label $l$. Intuitively, the ILP maximizes the semantic similarity of $K$ chosen exemplars, subject to each label $l$ being present at least once in the exemplars, and average prediction confidence of each data point for each label being greater than $\tau_l$.

Formally, the ILP is formulated as
\begin{align}
  \text{max} \sum_{i \in \mathcal{I} \setminus \{j\}} z_i &\cdot \text{sim}(q_i, q_j) \\
  \text{such that} \sum_{i \in \mathcal{I} \setminus \{j\}} z_i &= K \\
   z_{i} \cdot (\hat{y}^i_l - \tau_l)  \geq 0 \ \forall\ i &\in \mathcal{I} \setminus \{j\}, \forall\ l \in L_i \\
  \sum_{i \in \mathcal{I} \setminus \{j\}} z_{i} \cdot \text{count}(L_i, l) &\geq 1 \ \forall\ l \in \mathcal{L}
\end{align}

Here $\text{count}(L_i,l)$ denotes the number of occurrences of $l$ in $L_i$. 
In our experiments, we set $K=8$, and $\tau_l$ = $80^{th}$ percentile threshold of the set $\{\hat{y}^i_l\}^n_{i=1}$ for a particular label $l$. The idea is to have label-specific threshold since the fine-tuned model may not be equally calibrated for all labels. 

Since logits are not accessible for OpenAI LLMs GPT-3.5 and GPT-4x, in case Stage I labelling is done by either of these models using ICL, we skip the confidence thresholding constraint of ILP. This means that for this variant of SSP, the selection is made based on only similarity and label coverage.

\begin{table*}[h]
    \small
    \centering
    \begin{minipage}[t]{\textwidth}
    \begin{tabular}{llllll|l||lll|l}
         \textbf{Model} & \textbf{Hau} & \textbf{Ibo} & \textbf{Kin} & \textbf{Lug} & \textbf{Luo} & \textbf{Avg.} & \textbf{Fo} & \textbf{Got} & \textbf{Gsw} & \textbf{Avg}\\
         \hline 
            \textit{zero-labelled (0-CLT)} \\
            Full Fine-Tuning (FFT) & 49.9 & 54.9 & 55.4 & 56.3 & 40.2 & 51.3 & 77.6 & 17.8 & 62 & 52.5\\
           CPG \cite{ustun2020udapter} & 48.6 & 50.4 & 52.6 & 54.3 & 38.6 & 48.9 & 77.3 & 16.9 & 63.9 & 52.7\\
           ZGUL & 52.2 & 56 & 53.7 & 54.5 & 44.4 & 52.2 & 77.2 & 21.1 & 65 & 54.4\\
           ICL-Llama-2-70b & 64.3 & 61.2 & 59.2 & 60.1 & 47.3 & 58.4 & 79.1 & 36.0 & 71.8 & 62.3\\
           ICL-GPT-3.5-turbo & 54.5 & 69.2 & 57.8 & 63.7 & 46.4 & 58.3 & 81.2 & 37.9 & 72.2 & 63.8\\
           ICL-GPT-4x & 64.7 & 80.8 & 64.6 & 71.0 & 53.3 & 66.9  & 81.3 & 66.5 & 82.3 & 76.7\\
           \hline
           SSP(ICL)-llama-2-70b  & 57.6 & 62.6 & 56.0 & 57.6 & 43.1 & 55.4 & 78.5 & 37.9 & 73.5 & 63.3\\
           SSP(ICL)-GPT-3.5-turbo & 62.8 & 68.4 & 64.0 & 63.8 & 47.6 & 61.3 & 82.4 & 63.2 & 79.4 & 75.0\\
           SSP(ICL)-GPT-4x & 67.2 & 79.6 & 63.3 & 74.1 & 54.4 & 67.7 & 81.8 & \textbf{73.7} & 85.4 & \textbf{80.3}\\
           SSP(ZGUL)-Llama-2-70b & 68.4 & 58 & 56.1 & 54.7 & 42.3 & 55.9 & 79.9 & 39.9 & 72.9 & 64.2\\
           SSP(ZGUL)-GPT-3.5 & 61.1 & 68.9 & 62.1 & 67.1 & 51.4 & 62.1 & \textbf{82.8} & 67.5 & 77 & 75.8\\
           SSP(ZGUL)-GPT-4x & \textbf{72.5} & 79.8 & \textbf{71.4} & \textbf{77.4} & \textbf{55.1} & \textbf{71.2} & 82.2 & 71.5 & \textbf{85.6} & 79.8\\
           \hline
           w/o Conf. thresholding & 71.3 & \textbf{81.9} & 69.2 & 74.6 & 52.7 & 69.9 & 82.8 & 57 & 81.4 & 73.7\\
           w/o Label Coverage & 71.1 & 79.8 & \textbf{71.4} & \textbf{77.4} & \textbf{55.1} & 71 & 82.2 & 71.6 & \textbf{85.6} & 79.8\\
           w/o both (sim-based) & 70.3 & 81.8 & 68 & 74.8 & 51.9 & 69.4 & 82.4 & 55.8 & 82.3 & 73.5\\
           w/o ILP (Random) & 64.1 & 77.6 & 61.5 & 66.1 & 46.6 & 63.2 & 80.6 & 54.8 & 80.9 & 72.1\\
           \hline
           \hline
           
           \textit{Translate-train (0-CLT-T)} \\
           ZGUL &  72.5 & 68.5 & 67.9	& 65.5 & 47.3 & 64.3 & - & - & - & - \\
           ICL-GPT-4x & 68.7 & 78.1 & 58.7	& 76.3	& 53.8	& 67.1 & - & - & - & -\\
           SSP(ZGUL)-GPT-4x & 75.1 & 76.7	& \underline{72.3} & \underline{79.9} & 54.4 & 71.7 & - & - & - & -\\
           SSP(ICL)-GPT-4x & 69.9 & 79.8 & 60.6 & 74.7 & 53.8 & 67.8 & - & - & - & -\\
           \hline
           \textit{Translate-test (0-CLT-T)} \\
           Self-fusion (GPT-4x) \cite{chen2023better} & 68.4	& 68 & 58.8	& 66.5	& 39.7 & 60.3 & 83	& -	& 70 & -\\
           SSP(Self-fusion)-GPT-4x  & 70 & 78.6 & 64.6	& 77	& 51.3	& 68.3  & 83.7 & - & 83.9 & - \\
           \hline
           \hline
           \textit{Unlabeled data (0-CLT-U)} \\
           AfriBERTa \cite{ogueji-etal-2021-small} & 75.4 & 79.1	& 64.9 & 54.7 & 39.3 & 62.7 & - & - & - & -\\
           ZGUL++ \cite{rathore2023zgul} & \underline{78.5} & 68.9 & 62.5	& 66 & 50.2	& 65.2 & 81.5 &	18.7 & 80.4	& 60.2\\
           SSP(ZGUL++)-GPT-4x & 75.6 & \underline{84.7} & 70.3	& 75.4 & 54.6 & \underline{72.1} & \underline{83.9} & 71.7 & \underline{86} & \underline{80.5} \\
           \hline
           \hline
           \emph{Skyline (GPT-4x)} & \emph{75.5} & \emph{85.9} & \emph{70.7} & \emph{73.6} & \emph{67.2} & \emph{74.6} & \emph{93.5} & \emph{80.7} & \emph{89.9} & \emph{88}\\ 
    \end{tabular}
    \end{minipage}
    \caption{Micro-F1 scores for African NER (left) and Germanic POS (right). Best 0-CLT results are bolded while overall best results are underlined. Translate-train baselines could not be run for POS tagging due to absence of label-projection models for POS. However, Translate-test was possible as label-projection is performed using GPT-4 (Exception being Gothic, as it's translation is not supported in NLLB-200). Statistical significance of bold numbers (0-CLT comparison): McNemar p-value = 0.008 and 0.0004, respectively.}
    \label{tab:african_ner}
\end{table*}

\begin{table*}[h]
    \small
    \begin{minipage}{0.45\textwidth}
    \begin{tabular}{llll|l}
         \textbf{Model} & \textbf{Aym} & \textbf{Gn} & \textbf{Nah} & \textbf{Avg.} \\
         \hline 
           \emph{0-CLT} \\
           mDeBerta$^{100}$ & 34.9	& 43.9 & 48.9 & 42.6\\
           mDeBerta$^{FT}$ & 33.9 & 47	& 46.9 & 42.6\\
           ICL-GPT-3.5 & 38.2 & 41.7	& 35.3 & 38.4\\
           ICL-GPT-4 & 32.8 & 55.8 & 42.2	& 43.6\\
          \hline
           SSP(ICL)-GPT-3.5 & 38.4 & 38.8 & 43.2	& 40.1\\
           SSP(ICL)-GPT-4 &  37.5	& 58.5 & 51.8 & 49.3\\
           SSP(mDeBerta$^{FT}$)-Llama-2 & 36.5 & 37.8 & 41 & 38.4\\
           SSP(mDeBerta$^{FT}$)-GPT-3.5 & \textbf{43.1} & 46 & 46.8 & 45.3\\
           SSP(mDeBerta$^{FT}$)-GPT-4x & 36 & \textbf{61.3} & \textbf{59.2}	& \textbf{52.2}\\
            \bottomrule
            \end{tabular}
            
            \end{minipage} \hfill
            \begin{minipage}{0.45\textwidth}
            \begin{tabular}{llll|l}
            \toprule
            \textbf{Model} & \textbf{Aym} & \textbf{Gn} & \textbf{Nah*} & \textbf{Avg.} \\
            \midrule
           w/o Conf. & 42.9 & 60.1 & 50.3	& 51.1\\
           w/o Label& 37	& 58.2 & 57.4 & 50.9\\
           w/o both & 34.3	& 59.7 & 57.1 & 50.4\\
           w/o ILP (Random) & 33.4	& 53.8	& 53.4	& 46.9\\
           \hline
           \hline
           \textit{Translate Train} \\
           ICL-GPT-4 & 42.4 & 49.5 & - & - \\
           SSP(ICL)-GPT-4 & \underline{44.4} & 58.6 & - & - \\
           \hline
           \textit{Translate Test} \\
           ICL-GPT-4 & 36.4	& 45.5 & - & - \\
           SSP(ICL)-GPT-4 & 42.4	& 57.6 & - & -\\
           
           \hline
           \hline
           \emph{Skyline (GPT-4x)} & \emph{49.2} & \emph{55.6} & \emph{60} & \emph{54.9}\\
    \end{tabular}
\end{minipage}
\caption{Micro-F1 scores for Americas NLI (Statistical significance of bold number (0-CLT comparison): McNemar p-value = 0.054). * Nahuatl (Nah) not supported in NLLB-200.}
\label{tab:americas_nli}
\end{table*}

\section{Experiments}

Our main experiments assess SSP performance compared to existing state-of-the-art models for 0-CLT. We also wish to compare various SSP variants, and estimate the value of the ILP-based exemplar selection. 

\subsection{Tasks and Datasets}



We experiment on three tasks -- POS tagging, NER and Natural Language Inference (NLI). We use the UDPOS dataset \cite{nivre2020universal} for POS tagging over Germanic languages, MasakhaNER \cite{adelani2021masakhaner} for African NER, and AmericasNLI \cite{ebrahimi2022americasnli} for NLI task on the indigenous languages of Americas. Overall, we use eleven low-resource test languages as target (e.g., Kinyarwanda, Faroese, and Aymara), and 2-4 source languages per dataset (e.g., Icelandic, Spanish and Swahili; always including English). Further details are in Tables \ref{tab:source_count} and \ref{tab:target_count}.

Recent studies have shown sensitivity of the output to the template/format of input-output pairs written in the prompt \cite{Sclar2023QuantifyingLM, voronov2024mind}. We follow the best template given in \citet{Sclar2023QuantifyingLM} for NLI, while for sequence labelling, we explore various templates on our own and report our results on the best one. We refer to Appendix \ref{subsec: prompt} for details and the exact templates used for each of our tasks.

For obtaining test set, we randomly sample 100 test samples for each target language for NER and POS tasks. We justify this as each sentence has multiple labels, bringing the total no. of instances to be labeled per language to 2370 and 1100 for POS and NER respectively. For the NLI task, we sample 501 test samples (167 for each class: `entailment', `contradiction' and `neutral'). We report statistical significance (in table captions) to justify our evaluation.  

We also perform a careful contamination study, following \citep{ahuja-etal-2022-calibration}, by asking LLMs to fill dataset card, complete sentence (and labels), given partial sentence, and generate next few instances of the dataset. As further detailed in Appendix \ref{sec:contam}, we do not observe any evidence of contamination for these languages' test splits in the OpenAI LLMs.

\subsection{Comparison Models}
\textbf{Zero-shot Baselines:} We compare our SSP approach with the SoTA fine tuning models, as well as LLM-based ICL methods using naive random exemplar selection. In particular, we fine-tune ZGUL -- mBERT Language Adapter-based SoTA zero-shot baseline for NER and POS tagging, and mDeBERTa fine-tuned for NLI. We additionally utilize the public model mDeBERTa-v3-base-xnli \cite{laurer_less_2022} for NLI evaluation. We term our own fine-tuned model as mDeBERTa$^{FT}$ and the public model as mDeBERTa$^{100}$, as it was trained on 100 languages (excluding our target languages). For POS and NER, we also add full parameter fine-tuning and Conditional Parameter Generation (CPG \cite{ustun2020udapter}) baselines, all fine-tuned using the same underlying LM (i.e. mBERT). 


\vspace{0.5ex}
\noindent
\textbf{SSP Variants:} 
We implement SSP with a series of top-of-the-line LLMs -- GPT-3.5-turbo \cite{ouyang-etal-2022-impact}, GPT-4x (GPT-4/GPT-4-Turbo) \cite{achiam2023gpt}, and LLaMa-2-70b \cite{touvron2023llama}. If Stage I uses ICL, then the same LLM is used for both stages I and II. Alternatively, ZGUL and mDeberta based methods are also used in Stage I of SSP. 

To understand the value of the ILP, we perform three ablations on exemplar selection strategy -- (a) without confidence thresholding (for fine-tuned LM), (b) without label coverage and (c) without both, i.e. pure similarity-based. The ablations are conducted with the best performing underlying LLM i.e. GPT-4x. 

\vspace{0.5ex}
\noindent

\textbf{Leveraging Translation Models and Unlabeled Data:} For a comprehensive evaluation, we use the cross-lingual label projection models \textit{Codec} \cite{le2024constrained} for translate-train and \textit{Self-fusion} \cite{chen2023better} for translate-test baselines. More details are provided in Appendix \ref{subsec:tran}. \newline
Additionally, we leverage unlabeled data in the target language to establish a stronger baseline. We use the AfriBERTa encoder \cite{ogueji-etal-2021-small} for African languages and ZGUL++ \cite{rathore2023zgul}, which utilizes target Wikipedia data to pre-train a target language adapter, and fuses it with MRL adapters for fine-tuning on MRL data.

\vspace{0.5ex}
\noindent
\textbf{Skyline:} To understand the current performance gap due to lack of target language training data, we also implement a skyline utilizing the available data for target languages and perform \textit{few-shot in-language similarity-based} exemplar selection (using Ada-002) for \textit{in-language} ICL to the LLM. 
\section{Results and Analysis}
\label{sec:res}


We present the results for all tasks in Tables \ref{tab:african_ner},  and \ref{tab:americas_nli}. ICL-$X$ represents ICL over an LLM $X$ with source language exemplars. SSP($model$)-$X$ represents the use of $model$ for Stage I followed by LLM $X$ for Stage II. In case ICL is used in Stage I, then same LLM $X$ is used in both stages.

Analyzing the results, we first observe that all ICL-$X$ baselines perform much better than previous fine-tuning approaches for the 0-CLT task. This reaffirms the importance of studying and improving in-context learning over very large language models for our setting. 

Comparing among SSP variants, it is not surprising that GPT-4 performance supercedes GPT-3.5, which is much better than Llama2 70B. 
We next compare ICL baselines and SSP variants, when using the same LLM. We find that SSP's two stage workflow consistently outperforms ICL by significant margins. In fact, in-language exemplars with very noisy labels from stage 1 (E.g. for Got language with GPT-3.5-Turbo) perform quite well. These observations underscore the value of target language exemplars in ICL, even at the cost of label noise. 
Moreover, we compare SSP with Stage I via ICL over an LLM vs. via a fine-tuning baseline (ZGUL or mDeBerta). Fine-tuning baseline for Stage I has two benefits -- it is cheaper (due to no LLM calls in Stage I), and has prediction confidence that can allow ILP to select highly confident Stage II exemplars. Due to the latter, in two of the three language groups, the use of a fine-tuning baseline performs much better, and in the third group, it is marginally behind due to weaker performance in one language (Gothic). This happens because ZGUL has a particularly poor performance on this language, leading to much noisier labels in Stage II exemplars.

Finally, we experiment on SSP in 0-CLT-U (full target wikipedia) and 0-CLT-T (Translation model) settings, as shown in Table \ref{tab:african_ner}. We observe that the order of stage I performance is 0-CLT-T (translate-test) < 0-CLT < 0-CLT-T (translate-train) < 0-CLT-U, and same order of performance gets translated in stage II as well, while stage II performance being consistently better than stage 1 in all scenarios. This validates our hypothesis that SSP is effective under varying levels of noise in stage I labelings.

Overall, our best 0-CLT SSP solution uses a fine-tuning baseline (ZGUL or mDeBerta) for Stage I and GPT-4 for Stage II, using its novel ILP-based exemplar selection. It outperforms closest 0-CLT baselines by around 3 F1 pts, on average, establishing a new state of the art for zero-labelled cross-lingual transfer to low-resource languages. 
The best SSP reported 0-CLT results are statistically significant compared to the second best counterpart using McNemar's test (p-values in Tables 1 and 2 captions). 
We believe that our work is a significant advancement to the existing paradigm \cite{tanwar-etal-2023-multilingual, nambi2023breaking}, which is restricted to optimizing only 1 round of In-context learning. 




\subsection{Ablation Study}
We now discuss the results of removing ILP components in Stage II exemplar selection. Tables \ref{tab:african_ner},  and \ref{tab:americas_nli} (last four rows) report the impact of removing confidence thresholding constraint, label coverage constraint, both of these constraints (i.e., just using similarity) from the ILP. The final row removes ILP completely and presents results of random exemplars in Stage II. All these ablations are done on SSP with ZGUL/mDeBerta for Stage I, as only those output prediction probabilities.


 
\begin{figure*}
    \centering
\includegraphics[width=1.05\textwidth,height=0.15\textheight]{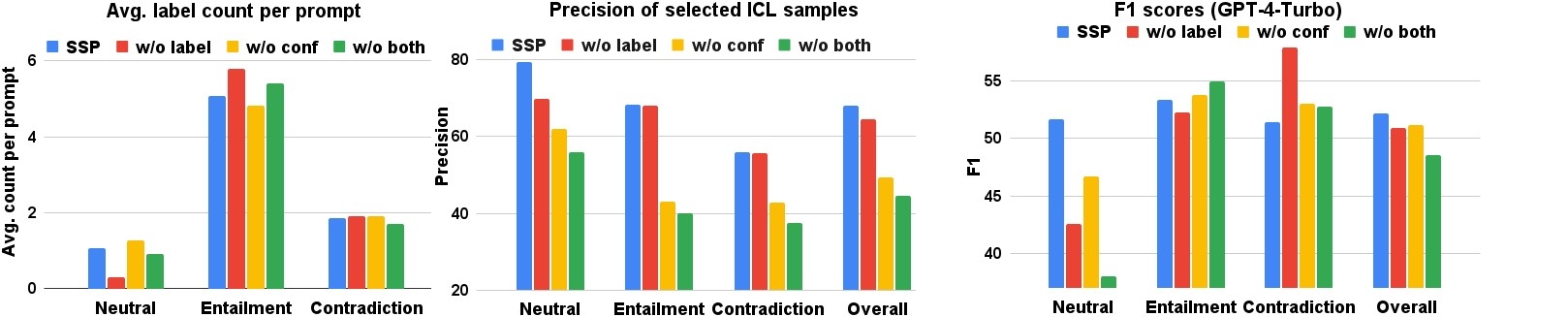}
    \caption{Label-wise statistics for AmericasNLI: Left to right - Label-wise count per prompt, Precision of ICL exemplars, and F1 results (averaged over target languages) using different selection strategies (GPT-4-Turbo)}
    \label{fig:lab_nli}
\end{figure*}

\textbf{Impact of label coverage:} We observe an average gain of 1.3 F1 points for AmericasNLI compared to the ablation model that does not impose label coverage constraint. We further compute the average number of exemplars for each label that are covered in the selected set for both methods, along with their label-wise F1 scores (see Figure \ref{fig:lab_nli}). We observe that the `neutral' label is not sampled in most cases for \textit{w/o label coverage} variant, while exactly one `neutral' label is sampled in the SSP(mDeBerta), with label constraint. This happens as the fine-tuned model mDeBerta-$FT$ has very poor recall (24) for `neutral' class and hence any selection strategy has a tendency to not sample this label, unless enforced via a constraint. The class-wise recall for SSP(DeBerta$^{CL}$)-GPT4 with and w/o label coverage are presented in Table \ref{tab:lab_nli_rec}. We observe a difference of 22 recall points for `neutral' class (57.6 vs 35.6) between the two ILP variants. An example illustrating this behavior is shown in Figure \ref{fig:neutral_error} (appendix). 
\begin{table}
    \centering
    \small
    \begin{tabular}{l|c|c|c|c}
         \textbf{Model} & \textbf{Neu.} & \textbf{Ent.} & \textbf{Con.} & \textbf{Macro-F1} \\
         \hline 
           DeBerta$^{CL}$ & 34.7 & 53 & 40.3 & 42.6\\
           SSP-V2 & \textbf{51.7} & \textbf{53.4} & 51.4	& \textbf{52.2}\\
           (w/o Label) & 42.6 & 52.3 & \textbf{57.9} & 50.9\\
    \end{tabular}
    \caption{Labelwise F1 scores for fine-tuned model (DeBerta-CL) and SSP-V2 variants w. and w/o Label coverage (GPT-4-Turbo)}
    \label{tab:lab_nli}
\end{table}

\begin{figure}
\includegraphics[width=0.5\textwidth]{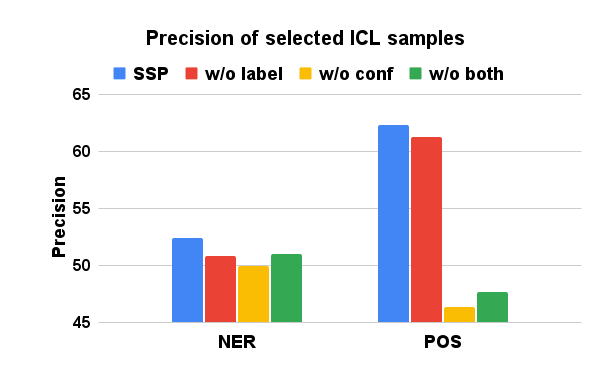}
    \caption{Precision of selected exemplars for African NER and Germanic POS}
    \label{fig:promp_ner_pos}
\end{figure}

\textbf{Impact of confidence thresholding:} 
For sequence labelling tasks, confidence thresholding plays a key role. This is validated from ablation results in Table \ref{tab:african_ner},  wherein removing confidence thresholding from SSP leads to 5.7 points drop for POS tagging (Germanic) and 1.3 points for NER. The drop is particularly significant (around 13.5 points) for Gothic (Got), which shows that not utilizing the confidence scores can lead to drastic drop. This may be because performance of ZGUL is already poor on Gothic (21 F1 points), but confidence thresholding may have likely compensated by picking higher quality exemplars. Removing thresholding would increase noise in exemplars considerably, leading to the drop (see figure \ref{fig:promp_ner_pos}). \newline
We further study its impact by computing the quality of Stage II exemplars selected by SSP(mDeBerta), as well as it's ablation variants. We compute the label-wise precision over all K$\times$N (K=8, N=no. of test instances) samples for each target language, and then report their macro-average. We observe for (Figure \ref{fig:lab_nli}) that the macro-precision of selected exemplars by full ILP is consistently higher than it's other ablation variants, the least value being of w/o both (similarity-based) variant. This implies that the ILP  is able to effectively sample high-precision (correctly labeled) exemplars which, in turn, gets translated into it's superior downstream performance on the task. \newline
For completeness, we also show the exemplar precision (correctness) statistics for NER and POS in Figure \ref{fig:promp_ner_pos}. The trends hold similar in the sense-that `w/o confidence' and `similarity-based' variants have significantly lower precision (higher noise) than SSP. This is expected because both these eschew confidence thresholding, leading to sampling of lower-confidence predictions. This translates to worse downstream performance (see Table \ref{tab:african_ner}). \newline 
We also note that w/o ILP (completely random selection) ablation performs much worse than SSP, showcasing the importance of carefully selecting the exemplar set. \newline
We present an error analysis of SSP approach in section \ref{error}.

\section{Conclusions and Future Work}
We study the zero-labelled cross-lingual transfer (0-CLT) setting for low-resource languages, when task-specific training data is available for related medium resource languages, along with unlabeled test data for target language. We present Self-Supervised Prompting (SSP) -- a novel two-stage framework for the use of in-context learning over very large language models. At a high-level, SSP first noisily labels the target test set using source training data (either by training a model/adapter) or by in-context learning over an LLM. SSP then uses these noisily labeled target data points as exemplars in in-context learning over the LLM. A key technical contribution is the use of integer-linear program that balances exemplar similarity, labelling confidence and label coverage to select the exemplars for a given test point. Thorough experiments on three NLP tasks, and eleven low-resource languages from three language groups show strongly improved performance over published baselines, obtaining a new state of the art in the setting. Ablations show the value each ILP component in downstream performance.

In the future, we seek to extend our technique to more non-trivial applications such as open generation tasks (E.g. summarization) and semantic parsing. We also posit that smaller fine-tuned models, when calibrated properly, can result in more efficient selection of exemplars to an LLM, as compared to poorly calibrated counterparts, in terms of downstream performance. We leave a careful and systematic investigation into this hypothesis for future work. 

\section{Limitations}
We show all our results and ablations on the recent state-of-the-art LLMs including GPT4. The inference for these LLMs is expensive, and makes the model deployment infeasible. Other potential limitations are extending our method to tasks such as fact checking, in which the LLMs suffer from \textit{hallucinations} and overprediction issues. The reason why we don't use LLM logits in ILP framework is because they are not openly released by OpenAI and hence, there becomes a need to rely on smaller fine-tuned models - which can potentially lead to sub-optimal downstream performance, in case the fine-tuned models are poorly calibrated. Another serious implication of using LLMs for non-roman script languages is unreasonably high \textit{fertility} (tokens per word split) of the LLM tokenizers, which increases the cost as well as strips the input prompt, which is not desirable.\newline
We also could not evaluate our approach on open generation tasks such as summarization, since their evaluation metrics are not reliable as to obtain a fair comparison of various models. Also, human evaluation could not be done at scale. That said, we note that every task is a generative task for LLM and we pose NLI as a short-form generation, while the POS and NER tasks as a templated long-form generation in current scope of our work. 

\bibliography{anthology,custom}

\begin{thebibliography}{40}
\expandafter\ifx\csname natexlab\endcsname\relax\def\natexlab#1{#1}\fi

\bibitem[{Achiam et~al.(2023)Achiam, Adler, Agarwal, Ahmad, Akkaya, Aleman, Almeida, Altenschmidt, Altman, Anadkat et~al.}]{achiam2023gpt}
Josh Achiam, Steven Adler, Sandhini Agarwal, Lama Ahmad, Ilge Akkaya, Florencia~Leoni Aleman, Diogo Almeida, Janko Altenschmidt, Sam Altman, Shyamal Anadkat, et~al. 2023.
\newblock Gpt-4 technical report.
\newblock \emph{arXiv preprint arXiv:2303.08774}.

\bibitem[{Adelani et~al.(2021)Adelani, Abbott, Neubig, D’souza, Kreutzer, Lignos, Palen-Michel, Buzaaba, Rijhwani, Ruder et~al.}]{adelani2021masakhaner}
David~Ifeoluwa Adelani, Jade Abbott, Graham Neubig, Daniel D’souza, Julia Kreutzer, Constantine Lignos, Chester Palen-Michel, Happy Buzaaba, Shruti Rijhwani, Sebastian Ruder, et~al. 2021.
\newblock Masakhaner: Named entity recognition for african languages.
\newblock \emph{Transactions of the Association for Computational Linguistics}, 9:1116--1131.

\bibitem[{Agrawal et~al.(2022)Agrawal, Zhou, Lewis, Zettlemoyer, and Ghazvininejad}]{Agrawal2022IncontextES}
Sweta Agrawal, Chunting Zhou, Mike Lewis, Luke Zettlemoyer, and Marjan Ghazvininejad. 2022.
\newblock \href {https://api.semanticscholar.org/CorpusID:254246450} {In-context examples selection for machine translation}.
\newblock In \emph{Annual Meeting of the Association for Computational Linguistics}.

\bibitem[{Ahuja et~al.(2023)Ahuja, Hada, Ochieng, Jain, Diddee, Maina, Ganu, Segal, Axmed, Bali et~al.}]{ahuja2023mega}
Kabir Ahuja, Rishav Hada, Millicent Ochieng, Prachi Jain, Harshita Diddee, Samuel Maina, Tanuja Ganu, Sameer Segal, Maxamed Axmed, Kalika Bali, et~al. 2023.
\newblock Mega: Multilingual evaluation of generative ai.
\newblock \emph{arXiv preprint arXiv:2303.12528}.

\bibitem[{Ahuja et~al.(2022)Ahuja, Sitaram, Dandapat, and Choudhury}]{ahuja-etal-2022-calibration}
Kabir Ahuja, Sunayana Sitaram, Sandipan Dandapat, and Monojit Choudhury. 2022.
\newblock \href {https://doi.org/10.18653/v1/2022.emnlp-main.290} {On the calibration of massively multilingual language models}.
\newblock In \emph{Proceedings of the 2022 Conference on Empirical Methods in Natural Language Processing}, pages 4310--4323, Abu Dhabi, United Arab Emirates. Association for Computational Linguistics.

\bibitem[{Alabi et~al.(2022)Alabi, Adelani, Mosbach, and Klakow}]{alabi2022adapting}
Jesujoba~O Alabi, David~Ifeoluwa Adelani, Marius Mosbach, and Dietrich Klakow. 2022.
\newblock Adapting pre-trained language models to african languages via multilingual adaptive fine-tuning.
\newblock In \emph{Proceedings of the 29th International Conference on Computational Linguistics}, pages 4336--4349.

\bibitem[{Asai et~al.(2023)Asai, Kudugunta, Yu, Blevins, Gonen, Reid, Tsvetkov, Ruder, and Hajishirzi}]{asai2023buffet}
Akari Asai, Sneha Kudugunta, Xinyan~Velocity Yu, Terra Blevins, Hila Gonen, Machel Reid, Yulia Tsvetkov, Sebastian Ruder, and Hannaneh Hajishirzi. 2023.
\newblock Buffet: Benchmarking large language models for few-shot cross-lingual transfer.
\newblock \emph{arXiv preprint arXiv:2305.14857}.

\bibitem[{Bergroth et~al.(2000)Bergroth, Hakonen, and Raita}]{lcs_algorithms}
L.~Bergroth, H.~Hakonen, and T.~Raita. 2000.
\newblock \href {https://doi.org/10.1109/SPIRE.2000.878178} {A survey of longest common subsequence algorithms}.
\newblock In \emph{Proceedings Seventh International Symposium on String Processing and Information Retrieval. SPIRE 2000}, pages 39--48.

\bibitem[{Brown et~al.(2020)Brown, Mann, Ryder, Subbiah, Kaplan, Dhariwal, Neelakantan, Shyam, Sastry, Askell et~al.}]{brown2020language}
Tom Brown, Benjamin Mann, Nick Ryder, Melanie Subbiah, Jared~D Kaplan, Prafulla Dhariwal, Arvind Neelakantan, Pranav Shyam, Girish Sastry, Amanda Askell, et~al. 2020.
\newblock Language models are few-shot learners.
\newblock \emph{Advances in neural information processing systems}, 33:1877--1901.

\bibitem[{Chen et~al.(2023{\natexlab{a}})Chen, Jiang, Ritter, and Xu}]{chen2023frustratingly}
Yang Chen, Chao Jiang, Alan Ritter, and Wei Xu. 2023{\natexlab{a}}.
\newblock Frustratingly easy label projection for cross-lingual transfer.
\newblock In \emph{Findings of the Association for Computational Linguistics: ACL 2023}, pages 5775--5796.

\bibitem[{Chen et~al.(2023{\natexlab{b}})Chen, Shah, and Ritter}]{chen2023better}
Yang Chen, Vedaant Shah, and Alan Ritter. 2023{\natexlab{b}}.
\newblock Better low-resource entity recognition through translation and annotation fusion.
\newblock \emph{arXiv preprint arXiv:2305.13582}.

\bibitem[{Chowdhery et~al.(2022)Chowdhery, Narang, Devlin, Bosma, Mishra, Roberts, Barham, Chung, Sutton, Gehrmann et~al.}]{chowdhery2022palm}
Aakanksha Chowdhery, Sharan Narang, Jacob Devlin, Maarten Bosma, Gaurav Mishra, Adam Roberts, Paul Barham, Hyung~Won Chung, Charles Sutton, Sebastian Gehrmann, et~al. 2022.
\newblock Palm: Scaling language modeling with pathways.
\newblock \emph{arXiv preprint arXiv:2204.02311}.

\bibitem[{Conneau et~al.(2020)Conneau, Khandelwal, Goyal, Chaudhary, Wenzek, Guzm{\'a}n, Grave, Ott, Zettlemoyer, and Stoyanov}]{conneau2020unsupervised}
Alexis Conneau, Kartikay Khandelwal, Naman Goyal, Vishrav Chaudhary, Guillaume Wenzek, Francisco Guzm{\'a}n, {\'E}douard Grave, Myle Ott, Luke Zettlemoyer, and Veselin Stoyanov. 2020.
\newblock Unsupervised cross-lingual representation learning at scale.
\newblock In \emph{Proceedings of the 58th Annual Meeting of the Association for Computational Linguistics}, pages 8440--8451.

\bibitem[{Devlin et~al.(2019)Devlin, Chang, Lee, and Toutanova}]{Devlin2019BERTPO}
Jacob Devlin, Ming-Wei Chang, Kenton Lee, and Kristina Toutanova. 2019.
\newblock \href {https://api.semanticscholar.org/CorpusID:52967399} {Bert: Pre-training of deep bidirectional transformers for language understanding}.
\newblock In \emph{North American Chapter of the Association for Computational Linguistics}.

\bibitem[{Ebrahimi et~al.(2022)Ebrahimi, Mager, Oncevay, Chaudhary, Chiruzzo, Fan, Ortega, Ramos, Gonzales, Meza-Ruiz et~al.}]{ebrahimi2022americasnli}
Abteen Ebrahimi, Manuel Mager, Arturo Oncevay, Vishrav Chaudhary, Luis Chiruzzo, Angela Fan, John Ortega, Ricardo Ramos, Annette~Rios Gonzales, Ivan Meza-Ruiz, et~al. 2022.
\newblock Americasnli: Evaluating zero-shot natural language understanding of pretrained multilingual models in truly low-resource languages.
\newblock In \emph{Proceedings of the 60th Annual Meeting of the Association for Computational Linguistics (Volume 1: Long Papers)}, pages 6279--6299.

\bibitem[{Garc{\'\i}a-Ferrero et~al.(2023)Garc{\'\i}a-Ferrero, Agerri, and Rigau}]{garcia2023t}
Iker Garc{\'\i}a-Ferrero, Rodrigo Agerri, and German Rigau. 2023.
\newblock T-projection: High quality annotation projection for sequence labeling tasks.
\newblock In \emph{Findings of the Association for Computational Linguistics: EMNLP 2023}, pages 15203--15217.

\bibitem[{He et~al.(2021)He, Liu, Gao, and Chen}]{he2021deberta}
Pengcheng He, Xiaodong Liu, Jianfeng Gao, and Weizhu Chen. 2021.
\newblock \href {https://openreview.net/forum?id=XPZIaotutsD} {Deberta: Decoding-enhanced bert with disentangled attention}.
\newblock In \emph{International Conference on Learning Representations}.

\bibitem[{Laurer et~al.(2022)Laurer, Atteveldt, Casas, and Welbers}]{laurer_less_2022}
Moritz Laurer, Wouter~van Atteveldt, Andreu~Salleras Casas, and Kasper Welbers. 2022.
\newblock \href {https://osf.io/74b8k} {Less {Annotating}, {More} {Classifying} – {Addressing} the {Data} {Scarcity} {Issue} of {Supervised} {Machine} {Learning} with {Deep} {Transfer} {Learning} and {BERT} - {NLI}}.
\newblock \emph{Preprint}.
\newblock Publisher: Open Science Framework.

\bibitem[{Le et~al.(2024)Le, Chen, Ritter, and Xu}]{le2024constrained}
Duong~Minh Le, Yang Chen, Alan Ritter, and Wei Xu. 2024.
\newblock \href {https://openreview.net/forum?id=DayPQKXaQk} {Constrained decoding for cross-lingual label projection}.
\newblock In \emph{The Twelfth International Conference on Learning Representations}.

\bibitem[{Min et~al.(2022)Min, Lyu, Holtzman, Artetxe, Lewis, Hajishirzi, and Zettlemoyer}]{min2022rethinking}
Sewon Min, Xinxi Lyu, Ari Holtzman, Mikel Artetxe, Mike Lewis, Hannaneh Hajishirzi, and Luke Zettlemoyer. 2022.
\newblock Rethinking the role of demonstrations: What makes in-context learning work?
\newblock In \emph{EMNLP}.

\bibitem[{Muller et~al.(2021)Muller, Anastasopoulos, Sagot, and Seddah}]{muller2021being}
Benjamin Muller, Antonios Anastasopoulos, Beno{\^\i}t Sagot, and Djam{\'e} Seddah. 2021.
\newblock When being unseen from mbert is just the beginning: Handling new languages with multilingual language models.
\newblock In \emph{Proceedings of the 2021 Conference of the North American Chapter of the Association for Computational Linguistics: Human Language Technologies}, pages 448--462.

\bibitem[{Nambi et~al.(2023)Nambi, Balloli, Ranjit, Ganu, Ahuja, Sitaram, and Bali}]{nambi2023breaking}
Akshay Nambi, Vaibhav Balloli, Mercy Ranjit, Tanuja Ganu, Kabir Ahuja, Sunayana Sitaram, and Kalika Bali. 2023.
\newblock Breaking language barriers with a leap: Learning strategies for polyglot llms.
\newblock \emph{arXiv preprint arXiv:2305.17740}.

\bibitem[{Nivre et~al.(2020)Nivre, de~Marneffe, Ginter, Hajic, Manning, Pyysalo, Schuster, Tyers, and Zeman}]{nivre2020universal}
Joakim Nivre, Marie-Catherine de~Marneffe, Filip Ginter, Jan Hajic, Christopher~D Manning, Sampo Pyysalo, Sebastian Schuster, Francis Tyers, and Daniel Zeman. 2020.
\newblock Universal dependencies v2: An evergrowing multilingual treebank collection.
\newblock In \emph{Proceedings of the Twelfth Language Resources and Evaluation Conference}, pages 4034--4043.

\bibitem[{{NLLB Team} et~al.(2022){NLLB Team}, Costa-jussà, Cross, Çelebi, Elbayad, Heafield, Heffernan, Kalbassi, Lam, Licht, Maillard, Sun, Wang, Wenzek, Youngblood, Akula, Barrault, Mejia-Gonzalez, Hansanti, Hoffman, Jarrett, Sadagopan, Rowe, Spruit, Tran, Andrews, Ayan, Bhosale, Edunov, Fan, Gao, Goswami, Guzmán, Koehn, Mourachko, Ropers, Saleem, Schwenk, and Wang}]{nllb2022}
{NLLB Team}, Marta~R. Costa-jussà, James Cross, Onur Çelebi, Maha Elbayad, Kenneth Heafield, Kevin Heffernan, Elahe Kalbassi, Janice Lam, Daniel Licht, Jean Maillard, Anna Sun, Skyler Wang, Guillaume Wenzek, Al~Youngblood, Bapi Akula, Loic Barrault, Gabriel Mejia-Gonzalez, Prangthip Hansanti, John Hoffman, Semarley Jarrett, Kaushik~Ram Sadagopan, Dirk Rowe, Shannon Spruit, Chau Tran, Pierre Andrews, Necip~Fazil Ayan, Shruti Bhosale, Sergey Edunov, Angela Fan, Cynthia Gao, Vedanuj Goswami, Francisco Guzmán, Philipp Koehn, Alexandre Mourachko, Christophe Ropers, Safiyyah Saleem, Holger Schwenk, and Jeff Wang. 2022.
\newblock No language left behind: Scaling human-centered machine translation.

\bibitem[{Ogueji et~al.(2021)Ogueji, Zhu, and Lin}]{ogueji-etal-2021-small}
Kelechi Ogueji, Yuxin Zhu, and Jimmy Lin. 2021.
\newblock \href {https://doi.org/10.18653/v1/2021.mrl-1.11} {Small data? no problem! exploring the viability of pretrained multilingual language models for low-resourced languages}.
\newblock In \emph{Proceedings of the 1st Workshop on Multilingual Representation Learning}, pages 116--126, Punta Cana, Dominican Republic. Association for Computational Linguistics.

\bibitem[{Ouyang et~al.(2022)Ouyang, Ye, and Li}]{ouyang-etal-2022-impact}
Siqi Ouyang, Rong Ye, and Lei Li. 2022.
\newblock \href {https://doi.org/10.18653/v1/2022.iwslt-1.9} {On the impact of noises in crowd-sourced data for speech translation}.
\newblock In \emph{Proceedings of the 19th International Conference on Spoken Language Translation (IWSLT 2022)}, pages 92--97, Dublin, Ireland (in-person and online). Association for Computational Linguistics.

\bibitem[{Pfeiffer et~al.(2020)Pfeiffer, Vuli{\'c}, Gurevych, and Ruder}]{pfeiffer2020mad}
Jonas Pfeiffer, Ivan Vuli{\'c}, Iryna Gurevych, and Sebastian Ruder. 2020.
\newblock Mad-x: An adapter-based framework for multi-task cross-lingual transfer.
\newblock In \emph{Proceedings of the 2020 Conference on Empirical Methods in Natural Language Processing (EMNLP)}, pages 7654--7673.

\bibitem[{Rathore et~al.(2023)Rathore, Dhingra, Singla et~al.}]{rathore2023zgul}
Vipul Rathore, Rajdeep Dhingra, Parag Singla, et~al. 2023.
\newblock Zgul: Zero-shot generalization to unseen languages using multi-source ensembling of language adapters.
\newblock In \emph{Proceedings of the 2023 Conference on Empirical Methods in Natural Language Processing}, pages 6969--6987.

\bibitem[{Reimers and Gurevych(2020)}]{reimers2020making}
Nils Reimers and Iryna Gurevych. 2020.
\newblock Making monolingual sentence embeddings multilingual using knowledge distillation.
\newblock In \emph{Proceedings of the 2020 Conference on Empirical Methods in Natural Language Processing (EMNLP)}, pages 4512--4525.

\bibitem[{Sclar et~al.(2023)Sclar, Choi, Tsvetkov, and Suhr}]{Sclar2023QuantifyingLM}
Melanie Sclar, Yejin Choi, Yulia Tsvetkov, and Alane Suhr. 2023.
\newblock \href {https://api.semanticscholar.org/CorpusID:264172710} {Quantifying language models' sensitivity to spurious features in prompt design or: How i learned to start worrying about prompt formatting}.
\newblock \emph{ArXiv}, abs/2310.11324.

\bibitem[{Tanwar et~al.(2023)Tanwar, Dutta, Borthakur, and Chakraborty}]{tanwar-etal-2023-multilingual}
Eshaan Tanwar, Subhabrata Dutta, Manish Borthakur, and Tanmoy Chakraborty. 2023.
\newblock \href {https://doi.org/10.18653/v1/2023.acl-long.346} {Multilingual {LLM}s are better cross-lingual in-context learners with alignment}.
\newblock In \emph{Proceedings of the 61st Annual Meeting of the Association for Computational Linguistics (Volume 1: Long Papers)}, pages 6292--6307, Toronto, Canada. Association for Computational Linguistics.

\bibitem[{Touvron et~al.(2023)Touvron, Martin, Stone, Albert, Almahairi, Babaei, Bashlykov, Batra, Bhargava, Bhosale et~al.}]{touvron2023llama}
Hugo Touvron, Louis Martin, Kevin Stone, Peter Albert, Amjad Almahairi, Yasmine Babaei, Nikolay Bashlykov, Soumya Batra, Prajjwal Bhargava, Shruti Bhosale, et~al. 2023.
\newblock Llama 2: Open foundation and fine-tuned chat models.
\newblock \emph{arXiv preprint arXiv:2307.09288}.

\bibitem[{{\"U}st{\"u}n et~al.(2020){\"U}st{\"u}n, Bisazza, Bouma, and van Noord}]{ustun2020udapter}
Ahmet {\"U}st{\"u}n, Arianna Bisazza, Gosse Bouma, and Gertjan van Noord. 2020.
\newblock Udapter: Language adaptation for truly universal dependency parsing.
\newblock In \emph{Proceedings of the 2020 Conference on Empirical Methods in Natural Language Processing (EMNLP)}, pages 2302--2315.

\bibitem[{Voronov et~al.(2024)Voronov, Wolf, and Ryabinin}]{voronov2024mind}
Anton Voronov, Lena Wolf, and Max Ryabinin. 2024.
\newblock Mind your format: Towards consistent evaluation of in-context learning improvements.
\newblock \emph{arXiv preprint arXiv:2401.06766}.

\bibitem[{Wan et~al.(2023{\natexlab{a}})Wan, Sun, Dai, Arik, and Pfister}]{wan-etal-2023-better}
Xingchen Wan, Ruoxi Sun, Hanjun Dai, Sercan Arik, and Tomas Pfister. 2023{\natexlab{a}}.
\newblock \href {https://doi.org/10.18653/v1/2023.findings-acl.216} {Better zero-shot reasoning with self-adaptive prompting}.
\newblock In \emph{Findings of the Association for Computational Linguistics: ACL 2023}, pages 3493--3514, Toronto, Canada. Association for Computational Linguistics.

\bibitem[{Wan et~al.(2023{\natexlab{b}})Wan, Sun, Nakhost, Dai, Eisenschlos, Arik, and Pfister}]{wan-etal-2023-universal}
Xingchen Wan, Ruoxi Sun, Hootan Nakhost, Hanjun Dai, Julian Eisenschlos, Sercan Arik, and Tomas Pfister. 2023{\natexlab{b}}.
\newblock \href {https://doi.org/10.18653/v1/2023.emnlp-main.461} {Universal self-adaptive prompting}.
\newblock In \emph{Proceedings of the 2023 Conference on Empirical Methods in Natural Language Processing}, pages 7437--7462, Singapore. Association for Computational Linguistics.

\bibitem[{Wei et~al.(2023)Wei, Wei, Tay, Tran, Webson, Lu, Chen, Liu, Huang, Zhou, and Ma}]{wei2023larger}
Jerry Wei, Jason Wei, Yi~Tay, Dustin Tran, Albert Webson, Yifeng Lu, Xinyun Chen, Hanxiao Liu, Da~Huang, Denny Zhou, and Tengyu Ma. 2023.
\newblock \href {http://arxiv.org/abs/2303.03846} {Larger language models do in-context learning differently}.

\bibitem[{Winata et~al.(2021)Winata, Madotto, Lin, Liu, Yosinski, and Fung}]{winata2021language}
Genta~Indra Winata, Andrea Madotto, Zhaojiang Lin, Rosanne Liu, Jason Yosinski, and Pascale Fung. 2021.
\newblock Language models are few-shot multilingual learners.
\newblock In \emph{Proceedings of the 1st Workshop on Multilingual Representation Learning}, pages 1--15.

\bibitem[{Zhang et~al.(2021)Zhang, Li, Chen, Deng, Bi, Tan, Huang, and Chen}]{zhang2021differentiable}
Ningyu Zhang, Luoqiu Li, Xiang Chen, Shumin Deng, Zhen Bi, Chuanqi Tan, Fei Huang, and Huajun Chen. 2021.
\newblock Differentiable prompt makes pre-trained language models better few-shot learners.
\newblock \emph{arXiv preprint arXiv:2108.13161}.

\bibitem[{Zhao et~al.(2021)Zhao, Wallace, Feng, Klein, and Singh}]{zhao2021calibrate}
Zihao Zhao, Eric Wallace, Shi Feng, Dan Klein, and Sameer Singh. 2021.
\newblock Calibrate before use: Improving few-shot performance of language models.
\newblock In \emph{International Conference on Machine Learning}, pages 12697--12706. PMLR.

\end{thebibliography}
\newpage
\appendix

\section{Implementation and Hyperparameter Details}
\label{sec:implement}
We use Azure OpenAI service \footnote{\href{https://azure.microsoft.com/en-in/products/ai-services/openai-service}{https://azure.microsoft.com/en-in/products/ai-services/openai-service}} for all experiments involving GPT-3x and GPT-4x models. For LLama-2-70b, we use the together API \footnote{\href{https://www.together.ai/}{https://www.together.ai/}}. We set temperature as 0.0 consistently for all our experiments, making our results directly reproducible. The max\_tokens (max. no. of generated tokens) parameter is set to 1024 for POS and NER tasks, while 15 for the NLI. For all experiments, the no. of exemplars ($M$) is fixed to 8 for uniform comparison. For ILP solver, we use Python's gurobipy \footnote{\href{https://pypi.org/project/gurobipy/}{https://pypi.org/project/gurobipy/}} package. The run-time for ILP per test query = 0.05 seconds, while that of pure similarity-based retrieval = 0.006 seconds. 

\subsection{Translation-based baselines}
\label{subsec:tran}
We explain both translate-train and translate-test methods as follows - 
\begin{itemize}
\item \textit{Translate-train}: Following \cite{le2024constrained}, we employ \textit{Codec} method to generate training data in target language X, $X^{train}$, using MRL labeled data. We perform stage 1 using following ways - 
\begin{enumerate}
    \item fine-tune a model on $X^{train}$, and infer on $X^{test}$
    \item perform ICL using exemplars from $X^{train}$ for each test query in $X^{test}$
\end{enumerate}
\item Following \cite{chen2023better}, we utilize \textit{Self-fusion} using GPT-4, that takes input as target query, it's English translation and English translation's annotations, appended as a prompt, and outputs the annotated target query.\footnote{We also tried Codec for translate-test, but could not reproduce the results reported in their paper for African languages (replicated avg. F1 = 60.5 v/s reported avg. F1 = 72).} 
\end{itemize}

\subsection{Estimating confidence $\hat{y}^i_k$}
\label{subsec:conf}
For NLI task, the model always predicts a single label: `neutral', `contradiction' or `entailment'. We simply apply softmax on the class logits for the predicted label to compute the confidence $\hat{y}^i_j$ (for $i^{th}$ test instance). \newline
In sequence labelling tasks, suppose for an input sentence having words: $\{w_1,w_2,...,w_T\}$, the model predicts labels $\{o_1,o_2,...,o_T\}$ with probabilities $\{\hat{p}_1,\hat{p}_2,...,\hat{p}_T\}$. Let $LabelSet$ be $\{l_{1}, l_{2}, ..., l_{N}\}$. We compute confidence $\hat{y}_l$ for each label for a given test example as follows: 
\begin{algorithmic}
\For{$k \gets 1$ to $N$}
    \State $\hat{y}_k \gets 0$ \Comment{init each label's confidence}
    \State $c_k \gets 0$ \Comment{init each label's count}
\EndFor
\For{$i \gets 1$ to $T$}
\For{$j \gets 1$ to $N$}
    \If{$l_{j}$ == $o_i$}
        \State $\hat{y}_j$ $\gets$ $\hat{y}_j + \hat{p}_i$  \Comment{Update $\hat{y}_{j}$}
        \State $c_j$ $\gets$ $c_j + 1$ \Comment{increase counter}
    \EndIf
\EndFor
\EndFor
\For{$k \gets 1$ to $N$}
\State $\hat{y}_k$ = $\hat{y}_k / c_k$  \Comment{average over all occurrences}
\EndFor
\end{algorithmic}
This outputs the confidence scores $\hat{y}_l$ for a given example, with those not predicted in a sequence having 0 value. 
\subsection{Dataset Details}
\begin{table}[h]
  \begin{center}
    \begin{tabular}{l|r|r}
      \textbf{Family} & \textbf{Source languages} & \textbf{Source size} \\
      \hline
      Germanic & \{En,Is,De\} & 30000 \\ 
      African & \{En,Am,Sw,Wo\} & 19788\\
      American & \{En,Es\} & 19998 \\
    \end{tabular}
  \caption{Size (No. of sentences) of Combined Source language datasets (En - English, Is - Icelandic, De - German, Am - Amharic, Sw - Swahili, Wo - Woloff, Es - Spanish)}
    \label{tab:source_count}
  \end{center}
\end{table}

\begin{table}[h]
  \begin{center}
    \begin{tabular}{l|r|r}
      \textbf{Family} & \textbf{Test languages} & \textbf{Labels} \\
      \hline
      Germanic & \{Fo, Got, Gsw\} & 2370\\ 
      African & \{Hau,Ibo,Kin,Lug,Luo\} & 1100\\
      American & \{Aym,Gn,Nah\} & 501\\
    \end{tabular}
  \caption{Size (No. of labels) of Target language datasets, \emph{per language}, on average. (Fo - Faroese, Got - Gothic, Gsw - Swiss German, Hau - Hausa, Ibo - Igbo, Kin - Kinyarwanda, Lug - Luganda, Luo - Luo, Aym - Aymara, Gn - Guarani, Nah - Nahuatl)}
    \label{tab:target_count}
  \end{center}
\end{table}
\section{Prompt details}
\label{subsec: prompt}

Prompts for the Named Entity Recognition (NER) and Part of Speech Tagging (POS) tasks are presented in the tab separated format shown in \ref{prompt:ner} and \ref{prompt:pos} respectively.

Prompts for Natural Language Inference (NLI) initially used the framework in \citet{ahuja2023mega} . To improve our performance, we changed the prompt to use \citet{Sclar2023QuantifyingLM}'s framework, where the authors performed an exhaustive search over tokens used for a prompt in order to find the prompt with optimal performance. This increased Macro F1 score by atleast 10\% across all the tested languages. We use the same prompt across all models used in our experiments.

\subsubsection{Natural Language Inference (NLI)}\label{prompt:nli}
\textbf{Task Description:} You are an NLP assistant whose purpose is to solve Natural Language Inference (NLI) problems. NLI is the task of determining the inference relation between two (short, ordered) texts: entailment, contradiction, or neutral. Answer as concisely as possible in the same format as the examples below:
\newline
\textbf{Input format:} \newline
Premise: \{premise\} , Hypothesis: \{hypothesis\} , \newline
\textbf{Output format:} \newline
Answer: \{output\} \newline
\textbf{Verbalizer: } \newline
match the one-word response from the model (neutral, contradiction or entailment)

\subsubsection{Named Entity Recognition (NER)}\label{prompt:ner}
\textbf{Task Description:} Tag the following sentence according to the BIO scheme for the NER task, using the tags PER (person), LOC (location), ORG (organization) and DATE (date). Follow the format specified in the examples below: \newline
\textbf{Input format:} \newline 
Sentence: $w_{1}\ w_2\ ...\ w_T$ \newline
\textbf{Output format:} \newline
Tags: \newline
$w_1$<TAB>$o_1$ \newline
$w_2$<TAB>$o_2$ \newline
... \newline
$w_T$<TAB>$o_T$ \newline
\textbf{Verbalizer: } \newline
Extract the sequence of labels $o_1, o_2, ... o_3$ from generated response.

\subsubsection{Part of Speech (PoS) tagging}\label{prompt:pos}
\textbf{Task Description:} Tag the following sentence according to the Part of Speech (POS) of each word. The valid tags are ADJ, ADP, ADV, AUX, CCONJ, DET, INTJ, NOUN, NUM, PART, PRON, PROPN, PUNCT, SCONJ, SYM, VERB, X. Follow the format specified in the examples below: \newline
\textbf{Input format:} \newline 
Sentence: $w_{1}\ w_2\ ...\ w_T$ \newline
\textbf{Output format:} \newline
Tags: \newline
$w_1$<TAB>$o_1$ \newline
$w_2$<TAB>$o_2$ \newline
... \newline
$w_T$<TAB>$o_T$ \newline
\textbf{Verbalizer: } \newline
Extract the sequence of labels $o_1, o_2, ... o_3$ from generated response.

\subsection{Verbalizer details for Tagging tasks}

The verbalizer for tagging tasks requires the LLM to output the words as well as the associated labels. The LLM's output may not be perfect, as it may fail to generate all words or associate a label with each word. As a result, we find the \textit{Longest Common Subsequence} between the words generated by the LLM and the words of the example. This is done using Dynamic Programming, as described in \cite{lcs_algorithms}.

Once we have found the longest common subsequence, we assign the corresponding tags generated by the LLM to these words. If the tags are invalid, we assign a default tag (O for NER, and X for POS). Finally, for the words which don't have any tags associated with them, we assign the same default tag as before.

It is to be noted that in most cases, the sentence generated by the LLM perfectly matches the original sentence. For GPT-4, less than 1\% of the words fell into the category of having an invalid tag generated, or not having the word generated.
\subsection{Error Analysis}
\label{error}
We investigate scenarios where SSP approach systematically fails compared to other methods. For NER, we find that ZGUL (fine-tuned LM) underpredicts the `DATE' label. As a result, SSP almost never samples this label in stage 2 exemplars, hence hurting the performance for this label. For NLI task, we observe that in order to ensure label coverage, SSP samples the underpredicted label `neutral' but while doing so, also ends up hurting the performance for `contradiction' label (as seen in last plot of Figure \ref{fig:lab_nli}). \newline

\subsection{Prompts for GSW Examples}
\begin{figure*}
    \includegraphics[height=2.3cm]{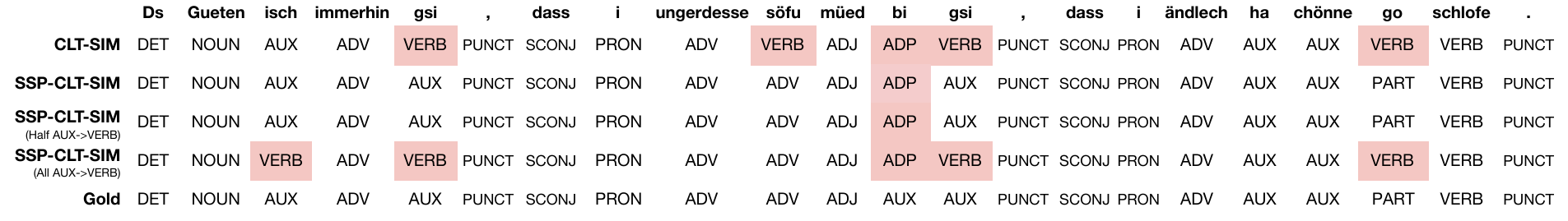}
    \includegraphics[height=2.3cm]{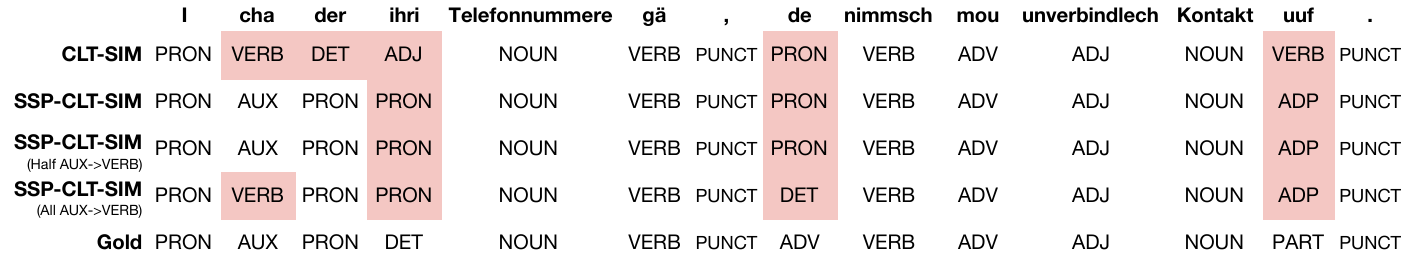}
    \caption{Label flips for CLT-SIM and SSP-SIM, for POS tagging in Swiss-German (gsw). Incorrect labels are marked in red. SSP-SIM ablations include flipping half/all of the AUX labels in the prompt to VERB labels. Gold labels are given for reference.}
    \label{fig:gsw_egs}
\end{figure*}

The base SSP-SIM prompts for the GSW examples highlighted in Figure \ref{fig:gsw_egs} are given below. Labels which are misclassified in the in-context exemplars are coloured in red, and the AUX labels which are to be flipped in the ablations are coloured in blue. It is interesting to note that examples 1 and 2 are similar, as example 1 is retrieved as an in-context exemplar for example 2.

\subsubsection{Example 1}

Tag the following sentence according to the Part of Speech (POS) of each word. The valid tags are ADJ, ADP, ADV, AUX, CCONJ, DET, INTJ, NOUN, NUM, PART, PRON, PROPN, PUNCT, SCONJ, SYM, VERB, X. Follow the format specified in the examples below:\\
Sentence: I main , das Ganze letscht Wuchä isch mier scho ächli iigfaarä .\\
Tags:\\
```\\
I	PRON\\
main	VERB\\
,	PUNCT\\
das	DET\\
Ganze	NOUN\\
letscht	ADJ\\
Wuchä	NOUN\\
isch	{\color{blue} AUX}\\
mier	PRON\\
scho	ADV\\
ächli	ADV\\
iigfaarä	VERB\\
.	PUNCT\\
```\\
Sentence: Du gsehsch uus , wi wenn de nöime no hättisch z trinken übercho .\\
Tags:\\
```\\
Du	PRON\\
gsehsch	VERB\\
uus	{\color{red} PRON}\\
,	PUNCT\\
wi	{\color{red} SCONJ}\\
wenn	SCONJ\\
de	{\color{red} DET}\\
nöime	{\color{red} ADJ}\\
no	ADV\\
hättisch	{\color{blue} AUX}\\
z	{\color{red} PART}\\
trinken	{\color{red} VERB}\\
übercho	VERB\\
.	PUNCT\\
```\\
Sentence: Dir weit mer doch nid verzöue , di Wäutsche heige vo eim Tag uf en anger ufghört Chuttlen ässe .\\
Tags:\\
```\\
Dir	PRON\\
weit	{\color{red} VERB}\\
mer	PRON\\
doch	ADV\\
nid	{\color{red} ADV}\\
verzöue	VERB\\
,	PUNCT\\
di	DET\\
Wäutsche	NOUN\\
heige	{\color{red} VERB}\\
vo	ADP\\
eim	DET\\
Tag	NOUN\\
uf	ADP\\
en	DET\\
anger	ADJ\\
ufghört	VERB\\
Chuttlen	NOUN\\
ässe	VERB\\
.	PUNCT\\
```\\
Sentence: es isch nämli echt usgstorbe gsi .\\
Tags:\\
```\\
es	PRON\\
isch	{\color{blue} AUX}\\
nämli	ADV\\
echt	{\color{red} ADJ}\\
usgstorbe	{\color{red} VERB}\\
gsi	{\color{blue} AUX}\\
.	PUNCT\\
```\\
Sentence: Aso bini rächt uufgschmissä gsi und dem entschprächend fascht verzwiiflät .\\
Tags:\\
```\\
Aso	ADV\\
bini	{\color{blue} AUX}\\
rächt	ADV\\
uufgschmissä	{\color{red} VERB}\\
gsi	{\color{blue} AUX}\\
und	CCONJ\\
dem	PRON\\
entschprächend	ADJ\\
fascht	ADV\\
verzwiiflät	VERB\\
.	PUNCT\\
```\\
Sentence: Der Ääschme wett nöd schaffe biin em .\\
Tags:\\
```\\
Der	DET\\
Ääschme	{\color{red} NOUN}\\
wett	{\color{blue} AUX}\\
nöd	{\color{red} ADV}\\
schaffe	VERB\\
biin	ADP\\
em	PRON\\
.	PUNCT\\
```\\
Sentence: Zerscht hends am Dani gsait , är söli dòch Hoochdütsch redä , das gängi denn grad gaar nöd , wenn är so redi , wiäner redi .\\
Tags:\\
```\\
Zerscht	ADV\\
hends	{\color{red} PRON}\\
am	ADP\\
Dani	PROPN\\
gsait	VERB\\
,	PUNCT\\
är	PRON\\
söli	{\color{blue} AUX}\\
dòch	ADV\\
Hoochdütsch	ADJ\\
redä	VERB\\
,	PUNCT\\
das	PRON\\
gängi	VERB\\
denn	ADV\\
grad	ADV\\
gaar	ADV\\
nöd	{\color{red} ADV}\\
,	PUNCT\\
wenn	SCONJ\\
är	PRON\\
so	ADV\\
redi	VERB\\
,	PUNCT\\
wiäner	{\color{red} PRON}\\
redi	VERB\\
.	PUNCT\\
```\\
Sentence: Isch das e Sach gsi , bis mer se gfunge hei gha .\\
Tags:\\
```\\
Isch	{\color{blue} AUX}\\
das	PRON\\
e	DET\\
Sach	NOUN\\
gsi	{\color{blue} AUX}\\
,	PUNCT\\
bis	SCONJ\\
mer	PRON\\
se	PRON\\
gfunge	VERB\\
hei	{\color{blue} AUX}\\
gha	{\color{red} VERB}\\
.	PUNCT\\
```\\
Sentence: Ds Gueten isch immerhin gsi , dass i ungerdesse söfu müed bi gsi , dass i ändlech ha chönne go schlofe .\\
Tags:\\
```\\

\subsubsection{Example 2}

Tag the following sentence according to the Part of Speech (POS) of each word. The valid tags are ADJ, ADP, ADV, AUX, CCONJ, DET, INTJ, NOUN, NUM, PART, PRON, PROPN, PUNCT, SCONJ, SYM, VERB, X. Follow the format specified in the examples below:\\
Sentence: I ha ar Marie-Claire gseit , es sig mer chli schlächt und i mög jetz nümm liire .\\
Tags:\\
```\\
I	PRON\\
ha	{\color{blue} AUX}\\
ar	{\color{red} PART}\\
Marie-Claire	PROPN\\
gseit	VERB\\
,	PUNCT\\
es	PRON\\
sig	{\color{blue} AUX}\\
mer	PRON\\
chli	ADV\\
schlächt	ADJ\\
und	CCONJ\\
i	PRON\\
mög	{\color{red} VERB}\\
jetz	ADV\\
nümm	{\color{red} ADV}\\
liire	VERB\\
.	PUNCT\\
```\\
Sentence: De Spanier hed de Kontakt vermettlet , d Rumäne sölled d Holländer ombrocht ha .\\
Tags:\\
```\\
De	DET\\
Spanier	NOUN\\
hed	{\color{blue} AUX}\\
de	DET\\
Kontakt	NOUN\\
vermettlet	VERB\\
,	PUNCT\\
d	DET\\
Rumäne	NOUN\\
sölled	{\color{blue} AUX}\\
d	DET\\
Holländer	{\color{red} PROPN}\\
ombrocht	VERB\\
ha	{\color{blue} AUX}\\
.	PUNCT\\
```\\
Sentence: Ds Gueten isch immerhin gsi , dass i ungerdesse söfu müed bi gsi , dass i ändlech ha chönne go schlofe .\\
Tags:\\
```\\
Ds	DET\\
Gueten	NOUN\\
isch	{\color{blue} AUX}\\
immerhin	ADV\\
gsi	{\color{red} VERB}\\
,	PUNCT\\
dass	SCONJ\\
i	PRON\\
ungerdesse	ADV\\
söfu	{\color{red} VERB}\\
müed	ADJ\\
bi	{\color{red} ADP}\\
gsi	{\color{red} VERB}\\
,	PUNCT\\
dass	SCONJ\\
i	PRON\\
ändlech	ADV\\
ha	{\color{blue} AUX}\\
chönne	{\color{blue} AUX}\\
go	{\color{red} VERB}\\
schlofe	VERB\\
.	PUNCT\\
```\\
Sentence: Isch das e Sach gsi , bis mer se gfunge hei gha .\\
Tags:\\
```\\
Isch	{\color{blue} AUX}\\
das	PRON\\
e	DET\\
Sach	NOUN\\
gsi	{\color{blue} AUX}\\
,	PUNCT\\
bis	SCONJ\\
mer	PRON\\
se	PRON\\
gfunge	VERB\\
hei	{\color{blue} AUX}\\
gha	{\color{red} VERB}\\
.	PUNCT\\
```\\
Sentence: De Dialäkt muess zu de Gschecht und zum Inhaut vonere Werbig passe .\\
Tags:\\
```\\
De	DET\\
Dialäkt	NOUN\\
muess	{\color{blue} AUX}\\
zu	ADP\\
de	DET\\
Gschecht	NOUN\\
und	CCONJ\\
zum	ADP\\
Inhaut	NOUN\\
vonere	ADP\\
Werbig	NOUN\\
passe	VERB\\
.	PUNCT\\
```\\
Sentence: Mit der Zit hani mi mit mir säuber uf ei Schriibwiis pro Wort aafo einige .\\
Tags:\\
```\\
Mit	ADP\\
der	DET\\
Zit	NOUN\\
hani	{\color{red} VERB}\\
mi	PRON\\
mit	ADP\\
mir	PRON\\
säuber	{\color{red} ADJ}\\
uf	ADP\\
ei	DET\\
Schriibwiis	NOUN\\
pro	ADP\\
Wort	NOUN\\
aafo	VERB\\
einige	{\color{red} DET}\\
.	PUNCT\\
```\\
Sentence: Mit all denä Wörter hani natürli nüt chönä aafangä .\\
Tags:\\
```\\
Mit	ADP\\
all	DET\\
denä	DET\\
Wörter	NOUN\\
hani	{\color{red} PRON}\\
natürli	ADV\\
nüt	{\color{red} ADV}\\
chönä	{\color{red} VERB}\\
aafangä	VERB\\
.	PUNCT\\
```\\
Sentence: Aso bini rächt uufgschmissä gsi und dem entschprächend fascht verzwiiflät .\\
Tags:\\
```\\
Aso	ADV\\
bini	{\color{blue} AUX}\\
rächt	ADV\\
uufgschmissä	{\color{red} VERB}\\
gsi	{\color{blue} AUX}\\
und	CCONJ\\
dem	PRON\\
entschprächend	ADJ\\
fascht	ADV\\
verzwiiflät	VERB\\
.	PUNCT\\
```\\
Sentence: I cha der ihri Telefonnummere gä , de nimmsch mou unverbindlech Kontakt uuf .\\
Tags:\\
```\\

\section{Source and Target Languages for each task}
\label{sec:lang}

\begin{table}[H]
    \label{tab:langcode}
    \begin{tabular}{l|r}
      \textbf{Code} & \textbf{Language}\\
      \hline
        En & English \\
        Am & Amharic \\
        Sw & Swahili \\
        Wo & Wolof \\
        Hau & Hausa \\
        Ibo & Igbo \\
        Kin & Kinyarwanda \\
        Lug & Luganda \\
        Luo & Luo \\
        Is & Icelandic \\
        De & German \\
        Fo & Faroese\\ 
        Got & Gothic \\ 
        Gsw & Swiss German \\
        Nds & Low-Saxon \\
        Es & Spanish \\
        Aym & Aymara \\
        Gn & Guarani \\
        Nah & Nahuatl \\
     \end{tabular}
    \caption{Languages and their codes}
\end{table}
\section{NLI Label coverage Analysis}
\begin{figure*}
\centering
\includegraphics[width=\textwidth]{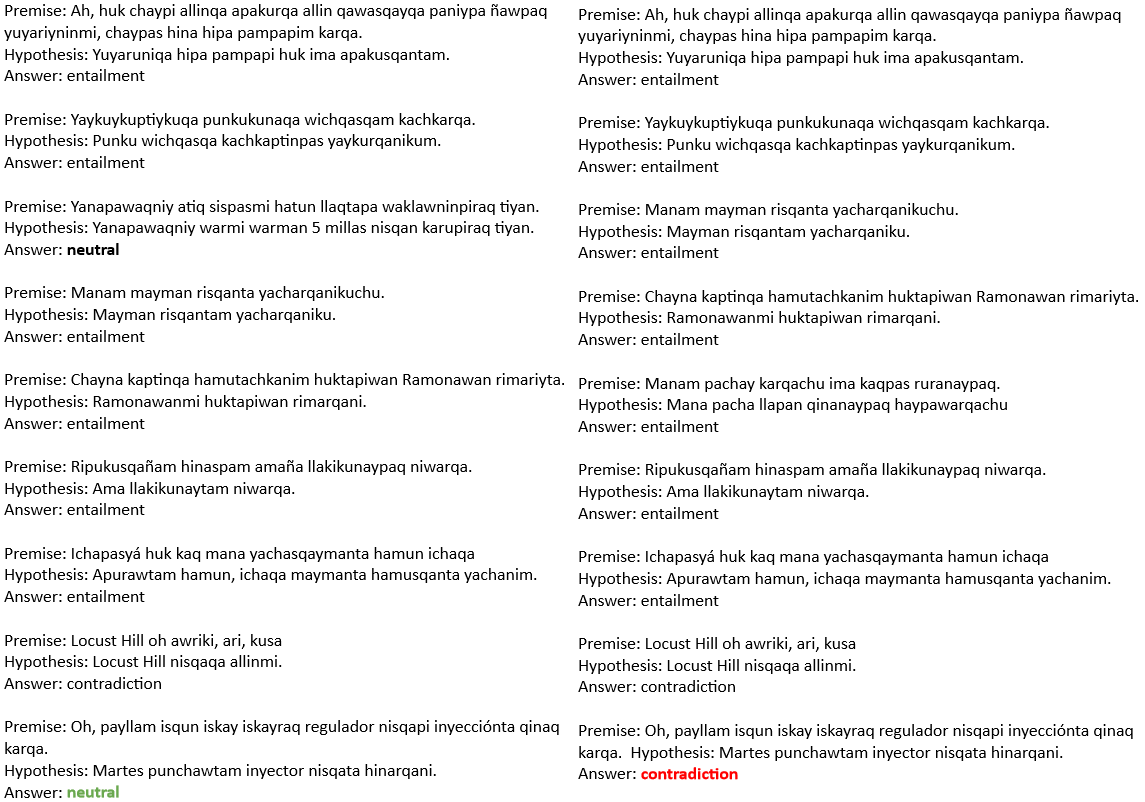}
    \caption{Correct case of `Neutral' detected by ILP (left), while `w/o label' variant misses it (right). We note that exact one `neutral' class has been sampled by ILP, while no `neutral' is sampled in `w/o label' version.}
    \label{fig:neutral_error}
\end{figure*}

\begin{figure*}
\centering
\includegraphics[width=\textwidth]{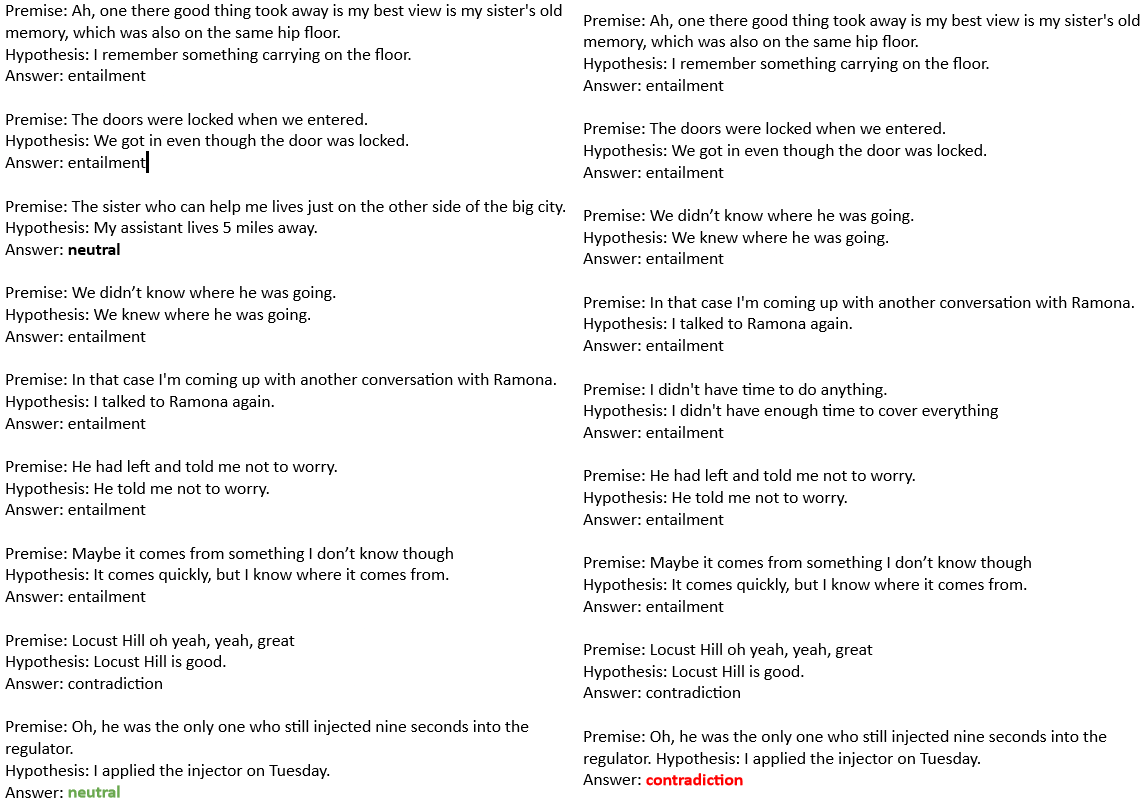}
    \caption{English translations of Exemplars shown in Fig. \ref{fig:neutral_error}}
    \label{fig:neutral_error1}
\end{figure*}

We present an example of correct prediction made by SSP as compared to the version that doesn't ensure label coverage in Figure \ref{fig:neutral_error} (English translation in Fig. \ref{fig:neutral_error1}).
\begin{table}
    \centering
    \begin{tabular}{l|c|c|c|c}
         \textbf{Model} & \textbf{Neu.} & \textbf{Ent.} & \textbf{Con.} & \textbf{Overall} \\
         \hline 
           DeBerta$^{CL}$ & 24.3 & \textbf{72.7} & 38.7 & 45.2\\
           SSP-V2 & \textbf{57.8} & 46.5 & 51.5 & 52\\
           (w/o Label) & 35.3 & 43.8 &	68.5 & 49.2\\
    \end{tabular}
    \caption{Labelwise Recall for fine-tuned model (DeBerta-based) and ILP variants w. and w/o Label coverage (GPT-4-Turbo)}
    \label{tab:lab_nli_rec}
\end{table}

\section{Qualitative Analysis: SSP-SIM}
\label{subsec:qual}
\begin{figure*}
    \centering
    \includegraphics[width=\columnwidth]{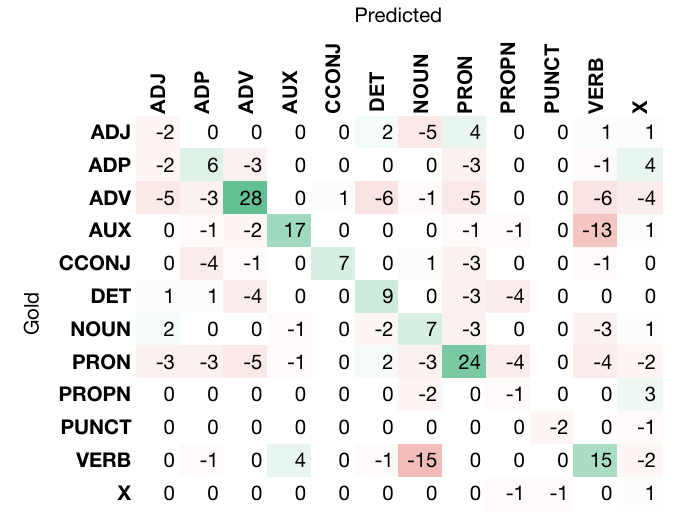}
    \caption{Difference in confusion matrices between similarity-based SSP Stage 1 and Stage 2 for the POS task, summed across all languages (tags with less than 100 instances have been omitted). The increase in correct tags is visible along the diagonal, and misclassifications between VERB and AUX tags / NOUN and VERB tags have also improved.}
    \label{fig:pos_improvements}
\end{figure*}

We present the analysis for the gains obtained via SSP-SIM for Germanic POS in Figure \ref{fig:pos_improvements}. The confusion matrix difference between SSP-SIM and CLT-SIM suggests that the model misclassifies auxiliary verbs as verbs in CLT-SIM, and this is corrected in SSP-SIM. These errors are a consequence of the labels on the in-context exemplars the model receives, and not the tokens of the language itself. 

We highlight this via the two Swiss-German POS examples in Figure \ref{fig:gsw_egs}. The misclassified verbs are corrected by SSP-SIM, and these labels are again misclassified when more than half of the labels in the in-context exemplars are corrupted.

\section{Data Contamination Analysis}
\label{sec:contam}

\begin{table*}
    \centering
    \begin{tabular}{l|>{\centering\arraybackslash}p{4cm}|>{\centering\arraybackslash}p{4cm}|>{\centering\arraybackslash}p{4cm}}
      \textbf{Task} & \textbf{Card Filling} & \textbf{Completion} & \textbf{Few-Shot Generation}\\
      \hline
      \raisebox{-2\height}{NER} & Didn’t predict correct languages; no split sizes generated & No match found & NA \\
      \hline
      \raisebox{-2\height}{POS} & predicted 33 languages, but doesn’t contain any of our target languages & No match found & NA \\
      \hline
      \raisebox{-2\height}{NLI} & predicts 3 languages, of which only one matches with our target language (Quechua); wrong test split size & Refuses to generate for 3 out of 4 target languages, except for Quechua - for which it predicts 100\% of the tokens wrong and only 40\% labels correctly (out of 10 instances) & Repeats the premise of last instance, copies the premise string to hypothesis as well (No match detected) \\
     \end{tabular}
    \label{tab:contam_study}
    \caption{Results of Contamination Study}
\end{table*}

Following Ahuja et al. 2023, we conduct contamination tests on test datasets for our target languages. We perform the following tests: 
\begin{itemize}
    \setlength{\itemsep}{-0.3em}
    \item Dataset Card filling: Generate dataset card (supported languages, dataset description, \#instances in each split, etc.)
    \item Completion: Given a few words, complete the sentence and their labels, and 
    \item Generation using first few instances: Given first K instances (K=5) in the dataset, generate next few instances following them.
\end{itemize}
We observe negligible contamination as depicted in table 8. The 40\% accuracy for Quechua was a result of all the labels passed for the exemplars being entailment labels. As a result, the model repeated the same label for all the other examples, giving a 40\% accuracy. \emph{Following these results, to prevent any chance of contamination, we remove Quechua from our evaluation dataset.}

\end{document}